\documentclass[letterpaper, 10 pt, conference]{ieeeconf}  

\IEEEoverridecommandlockouts                              

\usepackage{algorithm}
\usepackage{algcompatible}
\usepackage{algpseudocode}

\makeatletter
\let\OldStatex\Statex
\renewcommand{\Statex}[1][3]{%
  \setlength\@tempdima{\algorithmicindent}%
  \OldStatex\hskip\dimexpr#1\@tempdima\relax}
\makeatother
\usepackage{graphics} 
\usepackage{url}
\usepackage{booktabs}
\usepackage{circledtext}
\circledtextset{resize=real}

\usepackage{amsmath} 
\usepackage{amsfonts}
\usepackage{amssymb}  
\usepackage{multicol}
\usepackage{graphicx}
\usepackage{svg}
\usepackage{layouts}
\usepackage{wrapfig}
\usepackage{caption}
\usepackage{graphicx,subcaption}
\usepackage{easyReview}
\usepackage{printlen}
\usepackage{bm}
\usepackage{siunitx}
\usepackage{colortbl}	
\usepackage[utf8]{inputenc}
\usepackage{pgf}
\usepackage{tikz}
\usepackage{tikzscale}
\usepackage{pgfplots}
\DeclareUnicodeCharacter{2212}{\textendash}
\pgfplotsset{compat=1.14}
\usepgfplotslibrary{external}
\usepgfplotslibrary[external]
\tikzset{external/force remake}
\usepackage{algorithm}
\usepackage{algpseudocode}
\usepackage{makecell}
\usepackage{hyperref}

\title{\LARGE \bf
    Task and Motion Planning for Humanoid Loco-manipulation
}

\author{Michal Ciebielski$^{1}$, Victor Dh{é}din$^{1}$, Majid Khadiv$^{1}$ 
\thanks{This work was partially supported by the Huawei-TUM joint laboratory- Individual Project Agreements TC20231218038-2025-2 and TC20231218038-2025-3.}
\thanks{$^{1}$Munich Institute of Robotics and Machine Intelligence (MIRMI), Technical University of Munich (TUM), Germany. {\tt\small firstname.lastname@tum.de}}
\thanks{\tt\small $^{2}$\url{https://youtu.be/og3_T4VfXzc}}
}

\begin{document}

\maketitle
\thispagestyle{empty}
\pagestyle{empty}

\begin{abstract}
This work presents an optimization-based task and motion planning (TAMP) framework that unifies planning for locomotion and manipulation through a shared representation of contact modes. We define symbolic actions as contact mode changes, grounding high-level planning in low-level motion. This enables a unified search that spans task, contact, and motion planning while incorporating whole-body dynamics, as well as all constraints between the robot, the manipulated object, and the environment. Results on a humanoid platform show that our method can generate a broad range of physically consistent loco-manipulation behaviors over long action sequences requiring complex reasoning. To the best of our knowledge, this is the first work that enables the resolution of an integrated TAMP formulation with fully acyclic planning and whole body dynamics with actuation constraints for the humanoid loco-manipulation problem. A supplementary video demonstrating the results is available $^{2}$\href{https://youtu.be/og3_T4VfXzc}{here}.

\end{abstract}


\section{Introduction}
To perform any reasonable real-world task, a humanoid robot needs to plan whole-body loco-manipulation movements. 
Solving a holistic optimal control problem with contact constraints has been shown to successfully generate loco-manipulation behaviors in simulation \cite{tassa2012synthesis,mordatch2012discovery}. However, to achieve this these approaches relax complementarity constraints imposed by contact which introduces several artefacts and non-physical behaviors (e.g., force at a distance) in the generated trajectories, all the while being prone to poor local minima due to the non-convexity of the relaxed contact constraints. Furthermore, when multiple objects with sparse contact points (grabbing a handle to move an object or opening a door) are included, they struggle to find a solution.
Classical approaches for planning and control of loco-manipulation systems consider the effect of manipulated objects on the locomotion system as a disturbance \cite{penco2019multimode,thibault2022standardized,li2023multi}. However, for solving complex loco-manipulation problems, concurrent consideration of both locomotion and manipulation is essential. For instance, for picking an object off of a tall shelf, the robot may first need to move a table close to the shelf and stand on it in order to reach the object.


There have been recent efforts on the use of deep reinforcement learning (DRL) for loco-manipulation tasks in the real world \cite{ha2025learning}. However, these approaches are limited to very simple loco-manipulation tasks and cannot reason about more complex long-horizon goals. 
Imitation learning from tele-operation demonstration is another approach to performing complex loco-manipulation policies \cite{seo2023deep,liu2024opt2skill}. However, this approach requires expensive and laborious demonstration collection, which is in particular highly challenging for unstable loco-manipulation systems like humanoid robots.


To address long-horizon, sequential manipulation tasks, \cite{toussaint2018differentiable} used an optimization-based framework known as logic-geometric programming (LGP) \cite{toussaint2015logic}. In \cite{toussaint2018differentiable}, the integrated TAMP problem \cite{garrett2021integrated} was formulated as a tree search combining motion optimization with symbolic planning. To achieve this for sequential manipulation, a kinematic trajectory optimization augmented with logic predicates that constrain object dynamics was used. While the approach demonstrated strong performance on a stationary manipulator and later on a mobile manipulator \cite{hartmann2022long}, the omission of robot dynamics and robot-environment patch planning restricts the method's applicability to tasks where the robot remains statically stable and does not need to plan transitions between different patches in the environment.


To extend TAMP to quadrupedal loco-manipulation, \cite{sleiman2023versatile} adopted an optimization-based framework similar to that of \cite{toussaint2015logic, toussaint2018differentiable}. In particular, \cite{sleiman2023versatile} combined a first-order kinematics model augmented with full centroidal dynamics and complete object dynamics, enabling the generation of dynamically stable motions for a legged robot, which were successfully executed in real-world experiments. Despite these impressive results, the approach is constrained to motions with predefined locomotion cycles and a predefined number of end-effectors in contact with the ground. Moreover, tree expansion relies on reference generation and short-horizon optimization with only one full horizon refinement at the end, which restricts the diversity of achievable behaviors. Actuation limits are also not incorporated due to the use of a first-order kinematic model.


Compared to \cite{sleiman2023versatile}, we employ a second-order kinematic model with whole-body centroidal momentum dynamics, which allows us to incorporate actuation limits. This capability is crucial for executing dynamic motions on a humanoid robot, which is inherently unstable, and allows us to achieve fully acyclic planning for all end-effectors. Furthermore, in each tree expansion, we solve a full-horizon trajectory optimization, since adding a new constraint inevitably affects the preceding motion. In addition, unlike \cite{sleiman2023versatile}, which considers only discrete actions for manipulation, we unify the representation of locomotion and manipulation symbols, allowing us to reason over complex loco-manipulation skills. Finally, while \cite{sleiman2023versatile} demonstrates TAMP on quadrupeds with a single arm, we tackle the more challenging problem of humanoid loco-manipulation involving two arms.

\begin{figure*}
    \centering
    \includegraphics[width=0.9\textwidth]{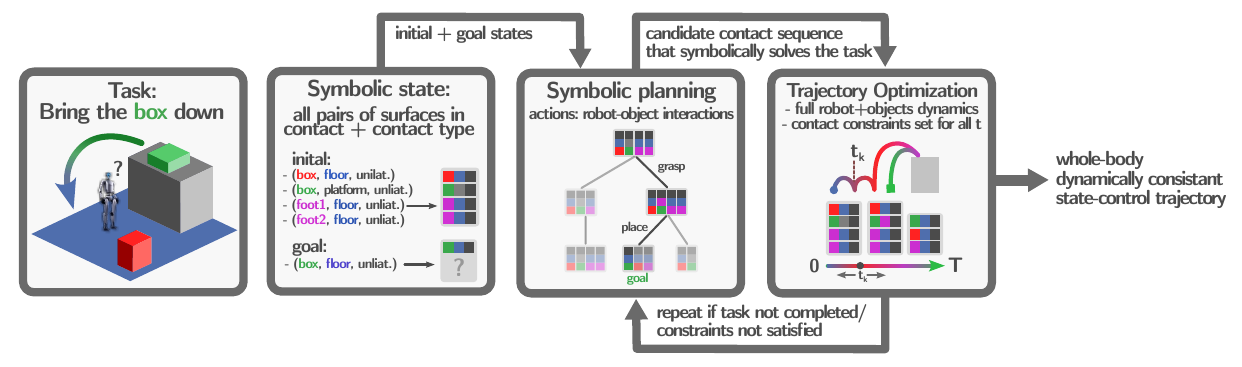}
    \vspace{-4mm}
    \caption{Overview of the proposed framework. \textbf{Second panel}: the task and the scene are translated into our symbolic framework at the level of contact. Our symbolic state includes all the interfaces in contact in the scene as well as the type of contact (Section \ref{sec:symbolic_states}). \textbf{Third panel}: given a symbolic goal, a tree search algorithm plans for a sequence of contact leading to the goal. The search is guided by object-centric manipulation logic (Section \ref{sec:symbolic_pruning}). \textbf{Fourth panel}: a contact-explicit TO solver optimizes for state and control (of both robot and objects) to achieve the candidate contact plan. Time duration of each contact phase is optimized (Section \ref{sec:to_formulation}). The interface between symbolic planning and TO is detailed in Section \ref{sec:solving_tamp}.
    }
    \vspace{-6mm}
    \label{fig:framework}
\end{figure*}

The main contributions of this paper are as follows:
\begin{itemize}
    \item We present a TAMP formulation for whole-body, long-horizon loco-manipulation that enables simultaneous decision-making over task, contact, and motions under whole-body constraints. Our generic formulation enables fully acyclic end-effector planning and dynamic patch transitions, while a unified treatment of locomotion and manipulation decision variables allows reasoning over complex loco-manipulation behaviors. 
    \item We develop a dedicated solver based on a combination of graph search and continuous trajectory optimization to generate long-horizon loco-manipulation behaviors with a mixture of different contact interaction modalities, such as one-handed full grasps and bi-manual loco-manipulation. To the best of our knowledge, this is the first work that enables the resolution of an integrated TAMP formulation with fully acyclic planning and whole body dynamics with actuation constraints for the humanoid loco-manipulation problem.
\end{itemize}


The rest of the paper is structured as follows. Section \ref{sec:preliminaries} presents the extension of \cite{toussaint2018differentiable} to floating-base systems. Section \ref{sec:method} outlines our problem formulation, highlighting the key details that enable the generation of long-horizon humanoid loco-manipulation plans. In Section \ref{Sec:results}, we present the results and conclude our findings in \ref{sec:conclusion}.

\section{Preliminaries}\label{sec:preliminaries}

\subsection{Floating-base dynamics}
Our TAMP formulation in this paper is an extension of LGP, which was originally developed for manipulation problems in \cite{toussaint2018differentiable}. When extending to a floating-base legged robot, it becomes essential to consider the under-actuated part of the dynamics to make sure the generated motion is consistent with the floating-base dynamics and does not lead the robot to fall. In particular, as it is shown in \cite{wieber2006holonomy}, the dynamics of a floating-base system can be split into the following equations
\begin{subequations}
    \begin{align}
        &M^a \dot v + b^a = \tau + \sum_{i \in \mathcal{C}} J_{c,i}^{a\top} \lambda_i,\label{eq:actuated_dynamics}\\
        &M^u \dot v + b^u =  \sum_{i \in \mathcal{C}} J_{c,i}^{u\top} \lambda_i,\label{eq:unactuated_dynamics}
    \end{align}
\end{subequations}
where $M$ and $b$ are the mass matrix and nonlinear terms of the dynamics, $v \in \mathbb{R}^{6+n}$ is the velocities in the tangent space of the configuration manifold $q \in \mathbb{SE}(3) \times \mathbb{R}^{n}$. $\tau \in \mathbb{R}^n$ is the vector of actuation torques, $\lambda_i \in \mathbb{R}^6$ is the contact wrench at the $i$th contact interaction port with its Jacobian denoted by $J_{c,i}$. The set $i \in \mathcal{C}$ specifies all active contacts. The superscripts $u$ and $a$ stand for underactuated and actuated. Compared to a manipulator dynamics that has only \eqref{eq:actuated_dynamics}, a floating-base legged robot also has an un-actuated part \eqref{eq:unactuated_dynamics} that needs to be satisfied to have dynamically consistent trajectories. 

The unactuated part of the dynamics can be written down as a function of the centroidal states (center of mass $c \in \mathbb{R}^3$ and momenta around it $h \in \mathbb{R}^6$) of the floating-base robot
\begin{align}
    & \dot{h} \;=\;
      \begin{bmatrix}
         mg + \sum_{i} f_{i} \\
         \sum_{i}
           (r_{c,i}-c)\times f_{i} + \kappa_{i}
      \end{bmatrix}
\end{align}
where $m$ is the total mass of the robot and $g$ is the gravity vector, $f \in \mathbb{R}^3$ and $\kappa \in \mathbb{R}^3$ are the force and moment components of the contact wrench, and $r_{c,i} \in \mathbb{R}^3$ is the point of action of the wrench.

Compared to the original LGP formulation and follow-up works \cite{toussaint2018differentiable, toussaint2015logic, hartmann2022long}, including an underactuated system increases the number of discrete variables that need to be considered in planning. While the original works considered the discrete actions associated with manipulation, we must also consider the discrete variables associated with locomotion. How this is done without introducing cyclic heuristics is discussed in the following section.

\section{Method}\label{sec:method}

Our method relies on two main components that can be seen in Fig. \ref{fig:framework}: a symbolic contact planner and a contact-explicit trajectory optimization (TO) solver. The contact planner outputs a candidate contact sequence (for both robot and objects) based on a symbolic representation of the scene and the task. The TO solver optimizes a state-control trajectory (including robot and objects states) that satisfies the constraints imposed by the candidate contact sequence. The contact planner is queried iteratively until the solver returns a trajectory that satisfies all constraints of a given sequence and fulfills the task goals. This procedure jointly optimizes the symbolic and continuous variables.


\subsection{Symbolic state and actions}\label{sec:symbolic_states}
Compared with prior work \cite{toussaint2018differentiable,sleiman2023versatile}, which restricts discrete symbols to manipulation, we additionally model discrete variables for non-manipulation contact switches. This enables the discovery of acyclic contact sequences for the feet during multi-contact locomotion, or opportunistic creation of extra supportive contacts (e.g., a hand contact to lift heavy objects).
Motivated by this, we define symbols at the level of contacts rather than object interaction modes as in \cite{toussaint2018differentiable,sleiman2023versatile}. In our formulation, the symbolic state (robot + objects) is the set of active contacts between entities (robot–environment, robot–object, object-environment and object–object).
Following this perspective, performing a task can be seen as executing a sequence of actions that would modify the full contact state of the scene by making some new contacts (e.g., placing an object on another or placing a foot on an object), or breaking existing ones (e.g., lifting an object or releasing an end-effector).

Building on the contact-level representation introduced above, we first formalize the notion of an \emph{interface}. We define an interface as any surface that can interact with another one through contact, i.e., an object surface (patch) can be an interface, as well as a gripper, a hand, or a foot. Physical objects (robot included) can have multiple interfaces that enable various interaction modes between them. Note that fixed environment patches are also interfaces with which the robot's end-effectors and objects can interact.
For a given scene, the set of available interfaces, denoted by $\mathcal{I}$, is predefined and remains unchanged during planning.
Interfaces also allow different types of contact, e.g., unilateral point-surface, bilateral point-surface, and unilateral surface-surface. $\mathcal{T}$ is the set of symbolic contact types between two interfaces. No contact interaction is denoted by $\emptyset \in \mathcal{T}$.
Following these notations, our symbolic state $s$ is defined as all the pairs of interfaces in contact and their respective contact type:
\begin{align}
    s := \{
            (i, j, \theta_{i, j})) |
            i, j \in \mathcal{I} \times \mathcal{I},
            \theta \in \mathcal{T},
            \theta_{i, j} \neq \emptyset
        \}.
\end{align}

The set of all possible states is denoted by $\mathcal{S}_{\mathcal{I}}$.
A symbolic action $a$ causes a transition ${s \overset{a}{\rightarrow} s'}$ from one contact state to another, which specifies how the full contact state changes by choosing which interfaces keep, break, or make contact with a (possibly new) interface. We also exclude action repetition that would result in the same contact state.
This search space quickly becomes intractable as the number of interfaces grows. Pruning methods to make the symbolic search tractable are detailed in Section \ref{sec:symbolic_pruning}.


\subsection{Optimization Formulation}\label{sec:to_formulation}
To formulate our optimization problem, we first let 
$x: [0,T] \to \mathcal{X}$ be the state trajectory and \(u: [0,T] \to \mathcal{U}\) be the control trajectory of the robot-object system with a floating base robot with $n$ joints and $m$ objects:
\begin{equation}
\begin{aligned}
    &x := (x^r, x^{o_1}, \dots x^{o_m}),
    &u := (u^r, u^{o_1}, \dots u^{o_m}).
\end{aligned}
\end{equation}
The robot states correspond to its configuration space $q^r \in \mathbb{SE}(3) \times \mathbb{R}^n$, velocity in the tangent space of configuration manifold $v^r \in \mathbb{R}^{n+6}$ and centroidal momentum $h^r \in \mathbb{R}^{6}$. The object states are described by their pose $q^o \in \mathbb{SE}(3)$ and spatial velocities $\mathcal{V}^o \in \mathbb{R}^{6}$
\begin{equation}
    x^r := (q^r, v^r, h^r), \quad x^{o_i} := (q^{o_i}, \mathcal{V}^{o_i}).
\end{equation}
The robot controls consist of its generalized accelerations $\dot v^r \in \mathbb{R}^{n+6}$ given its $n$ joints and the contact wrenches $\lambda_{ee}^r \in \mathbb{R}^6$ applied at each of the $n_{ee}$ robot end-effectors. The object controls include the spatial accelerations $\dot{\mathcal{V}}^{o_i} \in \mathbb{R}^6$ and object-environment wrenches $\mathcal{W}^{o_i}_{env} \in \mathbb{R}^6$
\begin{equation}
    u^r := (\dot v^r, \lambda_{ee_1}^r,  \dots, \lambda_{ee_{n_{ee}}}^r), \quad u^{o_i} := (\dot{\mathcal{V}}^{o_i}, \mathcal{W}^{o_i}_{env}).
\end{equation}
Given these definitions, our TAMP optimization can be written as 
\begin{equation}\label{eq:TAMP}
\begin{aligned}
\min_{\substack{x, u, s_{1:K}, \\ a_{1:K-1}, \bar T_{1:K}}} & \quad \sum_{i=0}^{N-1} \phi (x, u) \, +  \phi_N (x_N) + w_T\sum_{k=1}^{K} \bar T_k\\
\text{s.t.} \quad 
& x_0 = x _{init}, \ \  s_1 = s_{init}, \ \ x_N \in X_{goal}, \\
& h(x_N,s_K) = 0, \quad g(x_N,s_K) \leq 0, \\
&\forall k\in \{1,2,...,K-1\}:\\
& \quad s_{k+1} = \Gamma(s_{k}, a_{k}),\\
& \quad \bar T_{min}<\bar T_k< \bar T_{max},\\
& \forall i\in \{1,2,...,N-1\}:\\
& \quad h(x_i, u_i, s_{k(i)}) = 0, \\ & \quad g(x_i, u_i, s_{k(i)}) \leq 0.
\end{aligned}
\end{equation}
where $x$ denotes the continuous state trajectory, which is the concatenation of the robot state $x^r$ and the object states $\{x^{o_1},\dots,x^{o_m}\}$ and $u$ denotes the continuous control trajectory which is similarly a concatenation of the robot and object controls $u^r$ and $\{u^{o_1},\dots,u^{o_m}\}$. The variables $s,a$ are symbolic states and actions, which are discrete valued. The transition function $\Gamma$ maps the current symbolic state $s_k$ and symbolic action $a_k$ to the next symbolic state $s_{k+1}$. We choose a discrete transcription of the problem into $N$ time-steps partitioned into $K$ stages, where each $k$ spans $N_k$ time-steps with $\sum_{k=1}^K N_k = N$.

The path constraints $h$ and $g$ may be mode-dependent and vary as a function of $s_k$ (such as dynamics), or mode-independent, such as joint limits. Note that mode-independent constraints can be seen as a sub-category of mode-dependent ones; hence, for brevity, we only wrote the mode-dependent path constraints in \eqref{eq:TAMP}. Each symbolic state $s_k$ is active for $N_k$ timesteps (which may differ across stages), and the mapping $k \mapsto i$ in $s_{k(i)}$ ensures that the appropriate symbolic state is enforced at each timestep $i$. The symbolic state $s_k$ defines the full contact mode between the robot, object, and environment, and through this, the corresponding dynamic and kinematic constraints. In our formulation, the end time $T$, the number of symbolic states $K$ and each symbolic state duration $\bar T_k$ are all decision variables, optimized jointly with the continuous robot and object trajectories. In the following, we detail the path constraints of the problem.


\subsubsection{Robot Dynamics}
The robot dynamics and continuity constraints are enforced via the following:
\begin{subequations}\label{eq:constraint eqs}
    \begin{align}
    & q^r_{t+1} = q^r_{t} \oplus\ v^r_{t} \, \Delta t, \quad v^r_{t+1} = v^r_{t} + \dot{v}^r_{t} \, \Delta t, \label{eq:robot_integration}\\
    & h_{t+1} = h_t + \dot{h}_t  \Delta t, \quad \ h = A(q) v, \label{eq:momenta}\\
    & \dot{h} \;=\;
      \begin{bmatrix}
         mg + \sum_{i} f_{i} \\
         \sum_{i}
           (r_{i}-c)\times f_{i} + \kappa_{i}
      \end{bmatrix},\label{eq:CMD}\\
    &\tau_{\min} \preceq M^a \dot v + b^a - \sum_{i \in \mathcal{C}} J_{c,i}^{a\top} \lambda_i  \preceq \tau_{\max}. \label{eq:torque_constraint}
    \end{align}
\end{subequations}
We dropped the time dependence of the variables whenever necessary to avoid overloading the notation. The constraint in \eqref{eq:robot_integration} specifies the robot state integration, where $\oplus$ encodes a forward Euler update on the configuration manifold using the velocity $v_t$ over the timestep $\Delta t$.  The relationship between the robot velocities and centroidal momenta is encoded by \eqref{eq:momenta}, as well as the integration of centroidal momenta over time. Equation \eqref{eq:CMD} is the centroidal momentum dynamics. Finally, \eqref{eq:torque_constraint} applies the joint torque constraints ($\preceq$ is an element-wise inequality).

\begin{figure*}
    \vspace*{2mm}
    \centering
    \includegraphics[width=\linewidth]{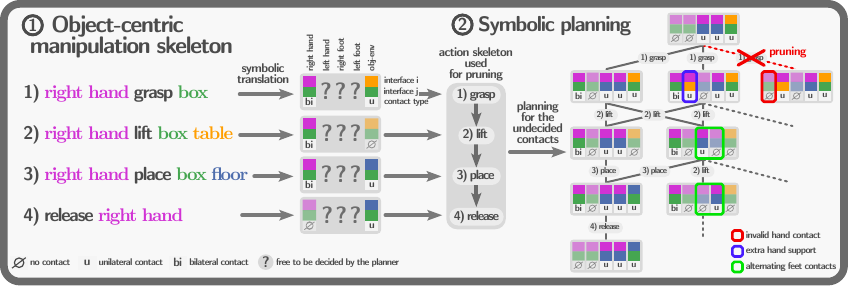}
    \caption{This figure illustrates how object-centric manipulation skeletons  are used to constrain the symbolic tree search (Only one skeleton is shown here, although multiple are used simultaneously in practice). \circledtext[height=2.ex,charshrink=0.65 ]{1} Each manipulation skeleton is first translated into the symbolic framework. This constrains some contact pairs while letting the others free to be decided by the planner (e.g., the feet and the left hand in this case, noted with "?"). \circledtext[height=2.ex,charshrink=0.65 ]{2} A minimal search tree is represented on the right side. Transitions that do not respect the contact pairs imposed by the skeleton are pruned (highlighted in red). Expanded states must satisfy the contact pairs imposed by the current \textbf{or} the next action of the skeleton. The left branch of the tree corresponds to a valid sequence that reaches the symbolic goal. The middle branch shows a repetition of the action \textit{lift}, which enables alternating feet contact (highlighted in green). Also in the middle branch, the first \textit{grasp} (highlighted in blue) is performed while the left hand adds a support contact on the table. Inactive interfaces are shown with the $\varnothing$ symbol.}
    \label{fig:symbolic-planning}
\end{figure*}
\subsubsection{Object Dynamics}
The following equations represent the object dynamics and state integration.
\begin{subequations}\label{eq:object_path_constr}
    \begin{align}
    & q^o_{t+1} = q^o_{t} \oplus\ \mathcal{V}^o_{t} \, \Delta t, \quad \mathcal{V}^o_{t+1} = \mathcal{V}^o_{t} + \dot{\mathcal{V}}^o_{t} \, \Delta t,\label{eq:object_integration}\\
    & \mathcal{W}^{o}_{env} + \mathcal{W}_{\text{grav}}^o + \sum_j \mathcal{W}_{j}^o = \mathcal{G}^o\,\dot{\mathcal{V}}^o - \bigl[\mathrm{ad}_{\mathcal{V}^o}\bigr]^\top \,\mathcal{G}_o\,\mathcal{V}^o\label{eq:twist_wrench_dynamics}
    \end{align}
\end{subequations}

Constraint \eqref{eq:object_integration} is the object state Euler integration and \eqref{eq:twist_wrench_dynamics} is the object twist-wrench dynamics as described in \cite{lynch2017modern}. $\mathcal{W}^o$ are wrenches expressed in the object frame, $\bigl[\mathrm{ad}_{\mathcal{V}^o}\bigr]$ is the lie bracket of $\mathcal{V}^o$ and $\mathcal{G}_o$ is the spatial inertia matrix. $\mathcal{W}_j$ represents the external wrenches at interfaces which are active with the object.
\subsubsection{Point-to-Patch Contact}  This type of contact is used to represent robot end-effectors modeled as points. Denoting the point index by $i$ and the patch index by $j$, we apply the following constraints:
\begin{subequations}\label{eq:point_to_patch_constr}
    \begin{align}
    & \left( {}^{j}\!r_{i} \right)_z = 0, \quad \left( {}^{j}\!v_{i}^{\text{rel}} \right)_{x,y} = 0 \label{eq:point_position},\\
    & \left|\left( {}^{j}\!r_{i} \right)_{x,y} \right| \leq {}^{j}\!\xi_{xy}  \quad \text{if } j \in \mathcal{J}_{\text{bnd}} \label{eq:point_position_bounds},\\
    & \left( {}^{j}\!R_i \, e_z \right)_z > 0, \quad \text{with } e_z = [0\ 0\ 1]^T \label{eq:halfspace_inequality},\\
    & \left.
    \begin{array}{l}
    \left\| \left( {}^{j}\!f_{i} \right)_{xy} \right\|_2 \leq \mu\, \left( {}^{j}\!f_{i} \right)_z \\
    \left( {}^{j}\!f_{i} \right)_z \geq 0, \quad {}^{j}\!\kappa_{i} = 0^3\\
    \end{array}
    \right\} \quad \text{if } \theta_{i,j} = \mathcal{T}_{\text{unilateral}}\label{eq:unilateral}.
    \end{align}
\end{subequations}
Equation \ref{eq:point_position} represents the position and velocity constraint for sticking contact. Note that in-plane positions $\left( {}^{j}\!r_{i} \right)_{xy}$ are free to be optimized. If patch $j$ is bounded, then constraint \eqref{eq:point_position_bounds} is active with symmetrical bounds $\xi_{xy}$. Furthermore, \eqref{eq:halfspace_inequality} constrains the limb of the end-effector to be in the positive halfspace of the patch in order to avoid limb-to-patch collision. Force constraints \eqref{eq:unilateral} are active if the contact type is unilateral. The wrench and velocity at point $i$ are represented in the patch frame $j$, i.e. ${}^{j}\!\mathcal{W}_i = [{}^{j}\!f_i, {}^{j}\!\kappa_i$], $\ {}^{j}\!\mathcal{V}_i = [{}^{j}\!v_i, {}^{j}\!\omega_i]$.

\subsubsection{Patch-to-Patch Contact}: This contact type occurs between end-effectors represented as patches (e.g. feet), patches on movable objects, and static environment patches as defined in Section \ref{sec:symbolic_states}. The following equations describe the constraints between patch $i$ and patch $j$: 
\begin{subequations}\label{eq:patch_to_patch_constr}
    \begin{align}
    & \left( {}^{j}\!r_{i} \right)_z = 0, \quad \left( {}^{j}\!v_{i}^{\text{rel}} \right)_{x,y} = 0 \label{eq:patch_patch_position},\\
    & \log_3 \left( {}^{j}\!R_i \right)_{x,y} = 0, \quad \left( {}^{j}\!\omega_{i}^{\text{rel}} \right)_{z} = 0 \label{eq:patch_patch_orientation},\\
    & \text{if } (i, j) \in \mathcal{J}_{\text{bnd}}: \quad \left|  \left( {}^{j}\!r_{i} \right)_{x,y}  \right|  \leq {}^{j}\!\xi_{xy} - {}^{j}\!R_i{}^{i}\!\xi_{xy} \label{eq:patch_patch_bounds},\\
    & \left.
    \begin{array}{l}
    \left\| \left( {}^{j}\!f_{i} \right)_{xy} \right\|_2 \leq \mu\, \left( {}^{j}\!f_{i} \right)_z, \\
    \left( {}^{j}\!f_{i} \right)_z \geq 0, \quad \ |\kappa_z| \le \mu_r\,{}^{i}\!f_z\,\\
    \left| \kappa_{xy} \right| \le {}^{i}\!f_z\,{}^{i}\!\xi_{xy},
    \end{array}
    \right\} \quad \text{if } \theta_{i,j} = \mathcal{T}_{\text{unilat.}}\label{eq:unilateral_patch}.
    \end{align}
\end{subequations}
In addition to the position and velocity constraints \eqref{eq:patch_patch_position}, the orientation is also constrained between patches such that their $xy$ planes are aligned \eqref{eq:patch_patch_orientation}, this is achieved by the matrix logarithm $\log_3$ which is applied on the rotation matrix ${}^{j}\!R_i$. Similarly to the point model, the position $\left( {}^{j}\!r_{i} \right)_{xy}$ and the in-plane yaw angle are free to be optimized. If both patches are bounded, i.e., neither of them is an unbounded environment patch, then \eqref{eq:patch_patch_bounds} is active. Furthermore, we model unilateral contact between patches with \eqref{eq:unilateral_patch}. In addition to the sticking friction coefficient $\mu$, we also include a torsional friction coefficient $\mu_r$.

\subsubsection{Collision Penalties}
In order to discourage the solver from optimizing paths that result in collision, we formulate the following penalties

\begin{subequations}\label{eq:penalties}
\begin{align}
& L_{\text{frame-plane}} = W \sigma(-{}^{P}\!p_{ee}^z) \label{eq:frame_plane} \\
& L_{\text{frame-box}} = W \prod_{i=1}^{3} \sigma({}^{B}\!H^i - |{}^{B}\!p_{ee}^i|) \label{eq:frame_box}\\
& L_{\text{box-box}} = W \prod_{i=1}^{3} \sigma({}^{B_1}\!H_{B_1}^i + {}^{B_1}\!H_{B_2}^i - |{}^{B_1}\!p_{B_2}^i|)\label{eq:box_box}
\end{align}
\end{subequations}

Each penalty uses a sigmoid $\sigma$ activation function over different collision penetration metrics. Equation \eqref{eq:frame_plane} measures the penetration of an end-effector with a plane $P$, while \ref{eq:frame_box} measures the penetration of an end-effector with a box $B$ with half-sizes $H$. Equation \ref{eq:box_box} measures the penetration of two boxes. We first express both boxes’ half-sizes in a common frame (that of $B_1$), sum them, and then compute the penalty using the centroid position of $B_2$.
In addition to using the box-box penalty for objects and the environment, the robot's torso and lower limbs (i.e. forearm/shin) are also approximated using a box, so collision penalties can be added between all elements of the scene.

\subsection{Solving TAMP Optimization}\label{sec:solving_tamp}

To solve the optimization problem in \eqref{eq:TAMP}, which includes both discrete and continuous decision variables, we combine symbolic graph search with gradient-based trajectory optimization. To achieve this, we proceed in two steps:
\begin{itemize}
    \item Symbolic planning: We search for a sequence of symbolic states and actions $(s_{1:K}, a_{1:K-1})$ that fulfills the task on the symbolic level.
    \item Trajectory optimization: For each symbolic plan, we impose the corresponding contact constraints and optimize for the continuous trajectory $(x, u)$.
\end{itemize}
This process is repeated over $N_{\text{it}}$ iterations, and we retain the trajectory with the lowest cost and constraint violation. 
It is also possible to rank the solutions based on a task completion criterion (e.g., the robot's terminal height if the task is to ascend as high as possible).

\subsubsection{Symbolic Planning}

We define a domain comprising all objects (static, movable, and robot) and their interfaces. The task is specified by an initial symbolic state $s_1 \in \mathcal{S}$ and a goal condition $s_g \in \mathcal{S}$.
From $s_1$, we build a search graph $\textsl{G}$ of all possible state transitions $(s \xrightarrow{a} s')$, limited to a maximum depth $K_{\text{max}}$ for tractability.
We search $\textsl{G}$ until we reach a state $s_K$ that satisfies the goal: $s_g \subseteq s_K$. Multiple such states may exist. For example, the goal \textit{box on table} could be satisfied with different foot contact configurations.

The search returns a sequence of valid contact transitions $s_1 \rightarrow \dots \rightarrow s_K$ that satisfies the symbolic goal of the task.
Note that any interfaces can be set in the goal condition, not just one of the objects. For example, a locomotion task can define target foot contacts. An empty goal condition is also valid. In that case, the search returns a feasible sequence of $K_{\text{max}}$ transitions that maximizes a secondary task objective such as base height maximization. 

\subsubsection{Dynamic Constraints}

Given the symbolic sequence, we set up the corresponding contact constraints for each phase in the trajectory optimization.
Specifically, for each symbolic phase $k \in \{1, \dots, K\}$ and optimization timestep $i \in \{0, \dots, N\}$, we enforce constraints on all active interface pairs $(i, j) \in \mathcal{I} \times \mathcal{I}$ with their contact types $\theta_{i,j}$, as specified in state $s_k$.
This results in a full set of constraints over the trajectory horizon. In practice, we assume each phase has a fixed number of optimization timesteps $N_k = N_s$, while the duration $\bar T_k$ of each phase is optimized.
\subsection{Symbolic search pruning}\label{sec:symbolic_pruning}
Pruning the symbolic search is essential to making the problem tractable. This is done at multiple levels as detailed below. 

\subsubsection{Infeasible states}

We reduced the size of $\mathcal{S}_{\mathcal{I}}$ by:
\begin{itemize}
    \item removing states with contact pairs between the same interface ($\theta_{i, i} \neq \emptyset$ is false).
    \item considering only one of the two symmetric contact pairs ($\theta_{i, j} \neq \emptyset \wedge \theta_{j, i} \neq \emptyset$ is false).
    \item removing states where two static interfaces are in contact with each other. For instance, different surfaces of the stairs cannot come into contact with each other.
\end{itemize}

\subsubsection{Infeasible transitions}

Transitions $(s \xrightarrow{a} s')$ where an interface $i$ is in contact with an interface $j$ in $s$ and $j' \neq j$ in $s'$ are pruned, as it is physically impossible for an interface to change its contact pair without breaking contact.

\subsubsection{Manipulation skeleton}

Since a static object can only be moved when the robot is in contact with it, the set of feasible transitions can be further reduced by adopting object-centric manipulation actions.
For example, if an object must be placed onto another interface, the contact sequence must, in this order, include transitions corresponding to the following high-level manipulation actions:

\begin{itemize}
    \item \textit{grasp}: make contact between one or more end-effectors and the object’s interface(s), resulting in either bilateral or unilateral constraints,
    \item \textit{lift}: remove the object-environment or object-object contact constraint by lifting the grasped object,
    \item \textit{place}: add a contact constraint between the object’s interface and a new supporting surface,
    \item \textit{release}: remove the contact constraint between the end-effector(s) and the object.
\end{itemize}

These actions fully define the contact sequence of both the manipulating end-effector(s) and the manipulated object’s interfaces.
Therefore, instead of exploring all possible transitions, we prune those that do not comply with a high-level, object-centric manipulation contact plan for the task—referred to here as a \textit{manipulation skeleton}. This skeleton can be generated using a logic-based planning framework such as PDDL.


In practice, we consider all skeletons symbolically satisfying the task in less than $N_{PDDL}$ actions during the search. Therefore, if two boxes have to be moved, both skeletons grasping box A first or box B first would be included in the search. We also plan for the skeletons first and then expand the tree to avoid expanding unnecessary states. A minimal example of symbolic planning with one object-centric manipulation skeleton can be seen in Fig. \ref{fig:symbolic-planning}.

\section{Results and discussion}\label{Sec:results}
In this section, we present the results of our method on two long-horizon loco-manipulation tasks. We first describe the tasks, then outline the implementation details, and finally present and discuss the results.

\subsection{Task Description}
We consider two distinct tasks (see Fig. \ref{fig:task_visualization}) that require reasoning over both locomotion and manipulation and solve them using the holistic optimization presented in section \ref{sec:solving_tamp}. To accomplish these tasks, the robot is allowed to choose end-effector contact sequences, robot and object patch interfaces, and different manipulation strategies.

\subsubsection{Platform Climbing}
\begin{figure}[h]
\centerline{\includegraphics[scale=0.12]{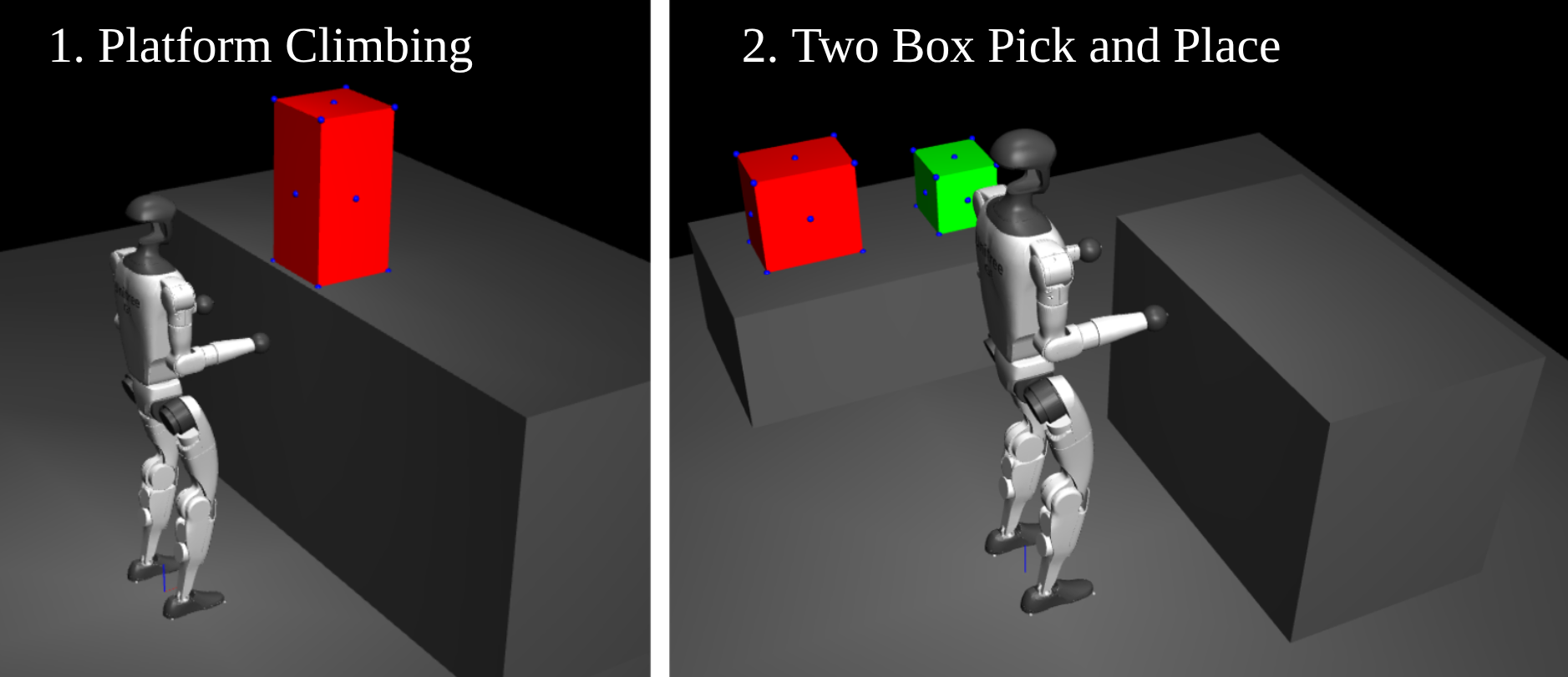}}
    \caption{In the platform climbing task, the goal is to ascend as high as possible. In the two-box pick and place task, the goal is to move the objects from the initial configuration to the empty table.}
    \label{fig:task_visualization}\vspace{-5mm}
\end{figure}

This task can be seen on the left side of Fig.~\ref{fig:task_visualization}; it has neither a predefined goal configuration nor a goal symbolic state for the robot or the object. Instead, the objective is for the robot to ascend as high as possible. This scenario highlights the importance of incorporating whole-body dynamics and actuation constraints in our framework, because without those, the trajectory optimization would result in high rewards for the sequence of directly jumping onto the platform. The task also illustrates a case in which unified locomotion and manipulation contact planning is necessary for task completion, due to the object dynamics influencing the stability of the robot. In this task, the red box (1 kg) features two opposite interfaces (patches) that allow for a bimanual grasp with unilateral contacts, and the robot is free to choose among different surface combinations to achieve this task. 
%
\subsubsection{Two Box Pick and Place}
A visualization of this task can be seen on the right side of Fig. \ref{fig:task_visualization}. In this task, the red box (1 kg) has one bilateral interface (i.e. a handle that the robot can grasp), while the green box (1 kg) has unilateral patch interfaces (i.e. only the bimanual grasping strategy is allowed). The goal configuration is for both boxes to be placed on the empty table as seen in the figure. The robot is free to choose the sequence of foot contacts as well as the order, sequence and type of manipulation. 

\subsection{Implementation Details}
Our trajectory optimization formulation was implemented using the Pinocchio \cite{carpentier2019pinocchio} library's integration with CasADI \cite{andersson2019casadi}. We then used IPOPT \cite{biegler2009large} to solve the resulting nonlinear program (NLP). All trajectories and scenes are visualized using the MuJoCo \cite{todorov2012mujoco} library.
We build the symbolic graph \textsl{G} mainly using the PDDL framework to define valid robot-object manipulation interactions at the symbolic level.
To explore this graph, we use Monte-Carlo tree search (MCTS), which simulates random action sequences and updates the graph based on the outcomes. 
In our case, symbolic plans that produce low trajectory constraint residuals and costs are rewarded. 
%
\begin{figure}[h]
\centerline{\includegraphics[scale=0.22]{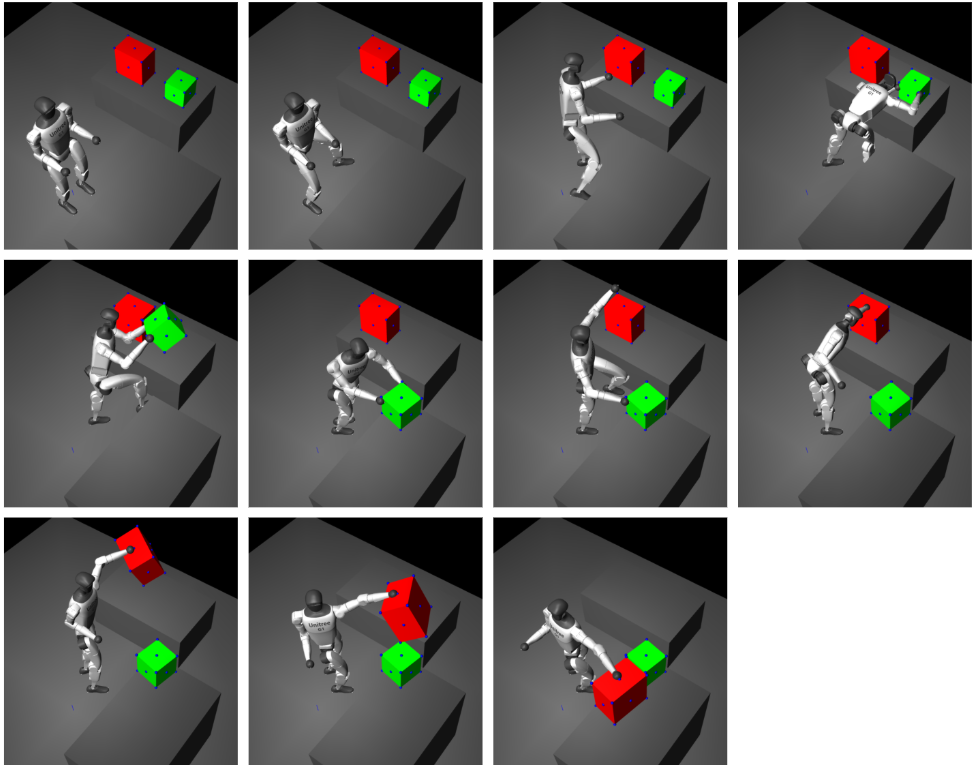}}
\caption{A solution to the two-box pick and place task. By leveraging the task-specific contact exploration, the solver is able to quickly find dynamically feasible solutions.}
    \label{fig:pick_and_place_solution}\vspace{-5mm}
\end{figure}
\subsection{Example Solutions}
The two-box pick and place tree search was expanded 200 times with an average solve time of 52.3 seconds. Note that using an NLP solver that exploits the sparsity in time of trajectory optimization, such as the SQP solver in Acados \cite{Verschueren2021acados}, could yield a significant speedup w.r.t. IPOPT. A variety of solutions were generated, with the first feasible solution satisfying the goal being found in 30 iterations. Figure \ref{fig:pick_and_place_solution} illustrates snapshots of the solution. More solutions with the top reward values that accomplish the task are presented in the accompanying video. This task highlights the benefit of combining task-specific goals with contact-patch selection: for example, in panel 7, the robot uses its free end-effector to stabilize itself while reaching for the red box.

\begin{figure}[H]
\centerline{\includegraphics[scale=0.29]{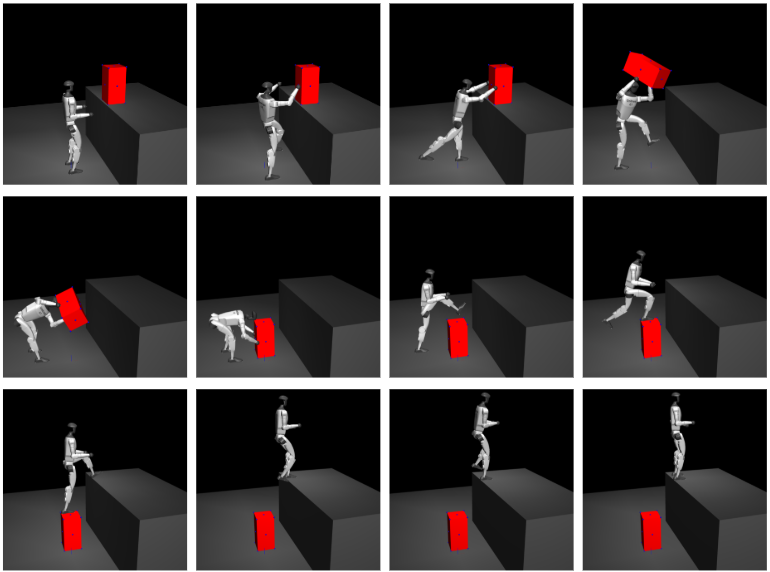}}
    \caption{A solution to the platform climbing task. Our formulation is able to accurately classify dynamically complex solutions thanks to the whole-body dynamics and actuation constraints embedded in our continuous optimization.}
    \label{fig:table_climb_solution}\vspace{-3mm}
\end{figure}

The platform-climbing task is significantly more challenging, since the tree search is driven by the continuous objective of maximizing the robot’s base height rather than by a discrete robot or object goal state. Moreover, most height-maximizing trajectories are dynamically infeasible. By embedding whole-body dynamics and actuation constraints into our optimization, we give the tree search a balanced signal that enforces both kinematic and dynamic feasibility alongside task-related rewards. Notably, all the highest-reward sequences use the box to climb onto the platform, demonstrating that, for long-horizon loco-manipulation tasks, incorporating whole-body dynamics and actuation constraints is crucial to effective exploration of the solution space of dynamically complex tasks. Figure \ref{fig:table_climb_solution} visualizes an example solution.
%

\begin{figure}[H]
    \centerline{\includegraphics[scale=0.25]{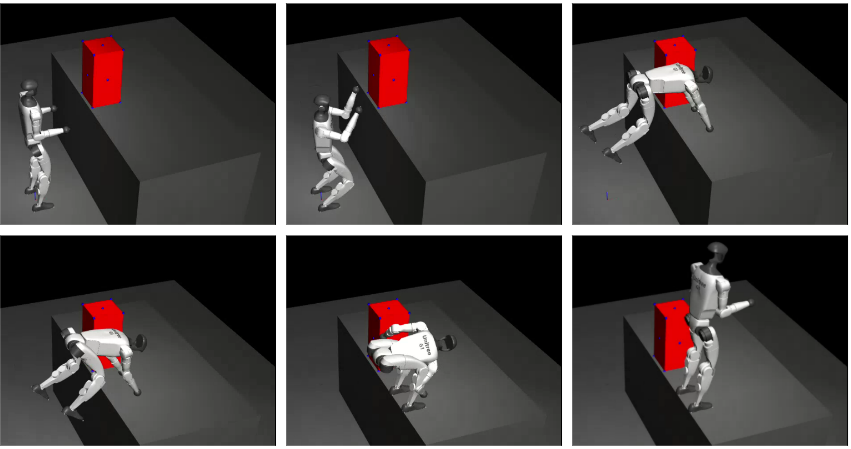}}
    \caption{Platform climbing task without torque constraints.}
    \label{fig:no_torque_jump}
\end{figure}

\begin{figure}[H]
    \centerline{\includegraphics[width=\linewidth]{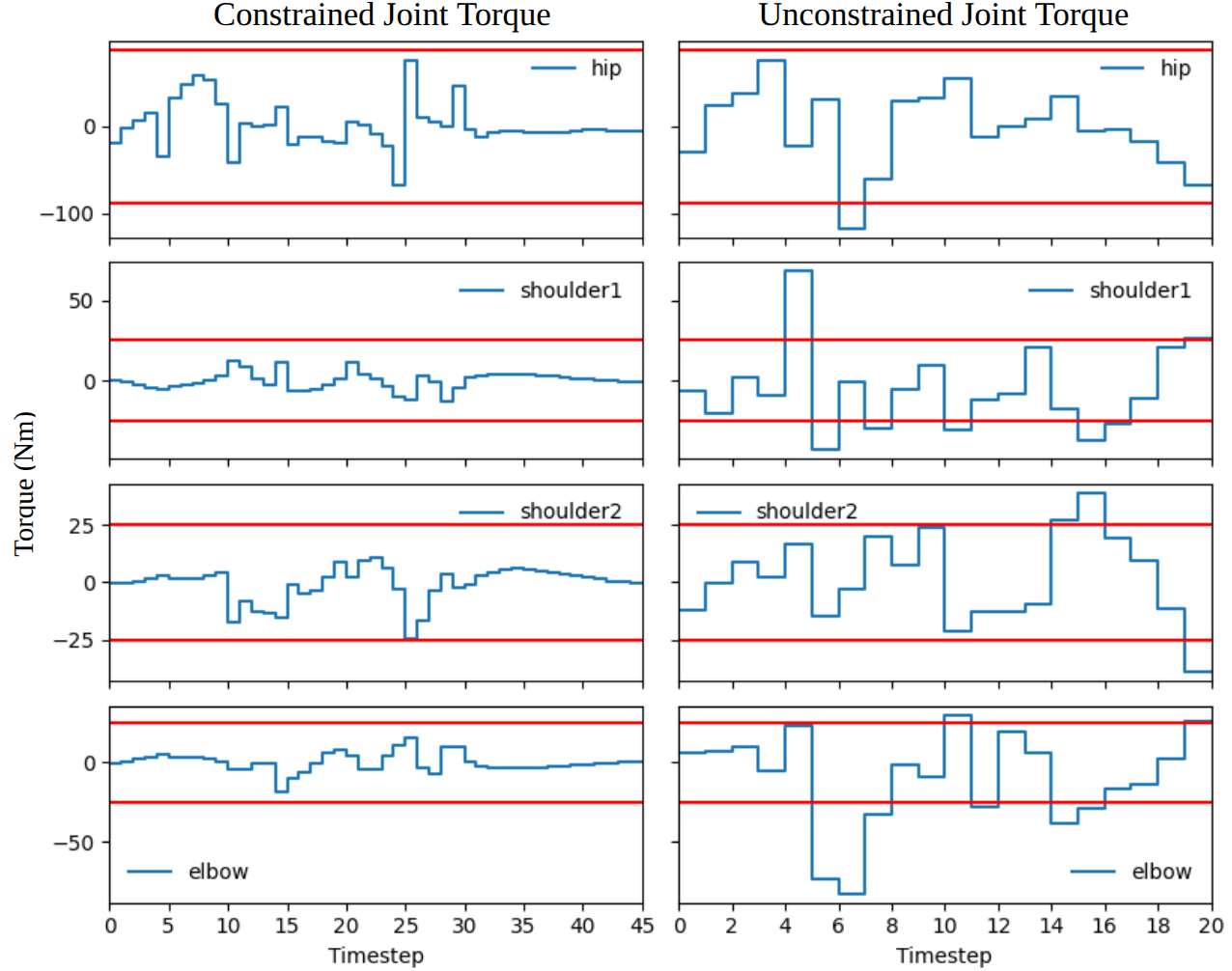}}
    \caption{Torque profiles for selected joints. \textbf{Right:} solution from a TAMP run without actuation constraints. \textbf{Left:} solution from our solver with actuation constraints enforced.}
    \label{fig:Torque_plots}
\end{figure}

To evaluate the impact of actuation constraints on the planner and the resulting behaviors, we reran the platform climbing task without enforcing the torque limits in~\eqref{eq:torque_constraint} (similar to \cite{sleiman2023versatile}). An example solution from this unconstrained TAMP run is shown in Fig.~\ref{fig:no_torque_jump}, where the robot jumps directly onto the table while using its end-effectors for stabilization (Fig. \ref{fig:no_torque_jump}). Two example solutions from this run are presented in the accompanying video. 
The torque profiles for the unconstrained solution are shown on the right of Fig.~\ref{fig:Torque_plots}, highlighting the joints with the largest violations. For comparison, the left side of Fig.~\ref{fig:Torque_plots} shows the torques for the same joints from a solution to the same task with actuation limits enforced.

\begin{figure}[H]
    \centerline{\includegraphics[width=\linewidth]{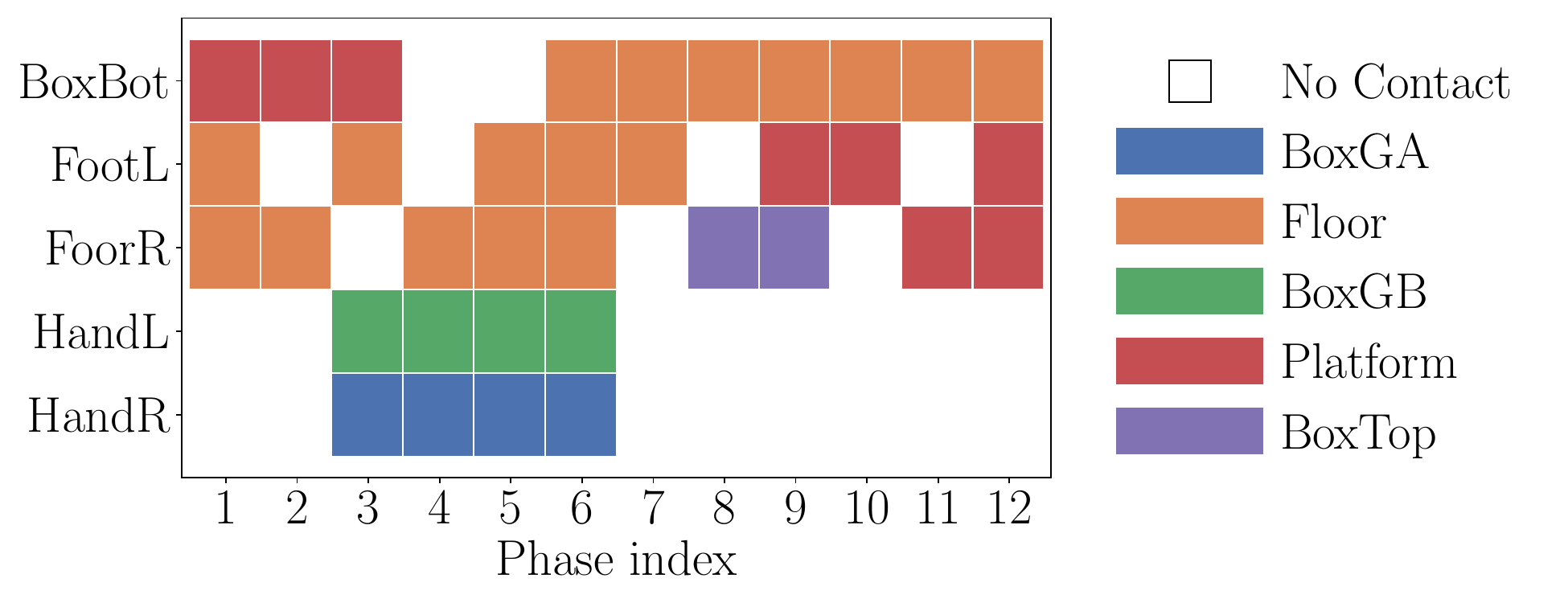}}\vspace{-3mm}
    \caption{Example of a contact sequence found by our framework for the platform climbing task. The sequence includes the interfaces of both the object and the robot. Note that there is no cyclic contact switch for the feet.}
    \label{fig:table_climb_sequence}
\end{figure}

In addition to analyzing the effect of actuation constraints, we also present an example of the acyclic output produced by our planner. Figure~\ref{fig:table_climb_sequence} depicts the acyclic, multi-patch contact mode sequence corresponding to the climbing task solution shown in Fig.~\ref{fig:table_climb_solution}.
The acyclic behavior illustrated in Fig.~\ref{fig:table_climb_sequence} is characteristic of solutions produced by our method. This is the result of a deliberate design choice: we avoid introducing cyclic heuristics that would, on the one hand, reduce the search space but, on the other hand, would strictly limit the diversity of possible behaviors. We emphasize generality and, through efficient pruning are able to constrain the discrete-variable tree search and thus solve such tasks as previously shown. Furthermore for dynamically challenging loco-manipulation tasks, especially on inherently unstable humanoid platforms, acyclic planning is essential, because predefining locomotion modes reduces the potential regions for stability and through this limits the range of tasks that can be completed.

\subsection{Discussion}
In addition to enabling acyclic behavior, our framework captures a wide range of behaviors by eliminating the need for alternative mechanisms to guarantee dynamic feasibility \cite{ponton2021efficient}, such as quasi-static assumptions \cite{tonneau2018efficient}, prescribed gait cycles \cite{dh2024diffusion,sleiman2023versatile}, or restrictions on the regions of the environment considered during planning \cite{toussaint2018differentiable}. While enforcing whole-body dynamics with actuation limits increases the complexity of the motion planning problem, this added complexity does not affect the size of the discrete search space. Moreover, simple heuristics that aim to guarantee dynamic feasibility often overconstrain the planner, limiting the diversity and adaptability of the resulting behaviors. Considering whole body dynamics with actuation limits is a key feature of this work and enabled the ability to plan dynamically feasible trajectories for a robot and objects simultaneously over multiple environment patches. 

\section{Conclusion}\label{sec:conclusion}

In this paper, we presented an optimization-based TAMP framework to reason about and generate long-horizon behaviors for humanoid loco-manipulation. In our proposed formulation, we defined symbolic actions that bring the system from one interaction mode to the next; hence, unifying locomotion and manipulation action predicates. This enabled us to reason over both locomotion and manipulation skills simultaneously. By combining graph search and gradient-based trajectory optimization, we developed a dedicated solver to solve the resulting mixed-integer optimization problem. We applied our framework to two challenging humanoid loco-manipulation scenarios, showcasing the capability of our framework to generate a long sequence of loco-manipulation behaviors automatically. In the future, we are planning to realize the behaviours in the real world.



\bibliography{master} 
\bibliographystyle{ieeetr}

\end{document}